\documentclass[letterpaper]{article} 
\usepackage{aaai24}  
\usepackage{times}  
\usepackage{helvet}  
\usepackage{courier}  
\usepackage[hyphens]{url}  
\usepackage{graphicx} 
\usepackage{natbib}  
\usepackage{caption} 
\usepackage{algorithm}
\usepackage{algorithmic}

\usepackage{newfloat}
\usepackage{listings}
\DeclareCaptionStyle{ruled}{labelfont=normalfont,labelsep=colon,strut=off} 
\floatstyle{ruled}
\newfloat{listing}{tb}{lst}{}
\floatname{listing}{Listing}
\nocopyright

\title{Identity-Aware Semi-Supervised Learning for\\Comic Character Re-Identification}
\author {
    Gürkan Soykan\textsuperscript{\rm 1,2},
    Deniz Yuret\textsuperscript{\rm 1,2}\equalcontrib,
    Tevfik Metin Sezgin\textsuperscript{\rm 1,2}\equalcontrib
}
\affiliations {
    \textsuperscript{\rm 1} Ko\c{c} University, KUIS AI Center\\
    \textsuperscript{\rm 2}Ko\c{c} University, Computer Engineering Department\\
    gsoykan20@ku.edu.tr, dyuret@ku.edu.tr, mtsezgin@ku.edu.tr
}

\usepackage{bibentry}

\usepackage{amsmath,amssymb}
\usepackage{booktabs}

\begin{document}

\maketitle

\begin{abstract}
    Character re-identification, recognizing characters consistently across different panels in comics, presents significant challenges due to limited annotated data and complex variations in character appearances. To tackle this issue, we introduce a robust semi-supervised framework that combines metric learning with a novel 'Identity-Aware' self-supervision method by contrastive learning of face and body pairs of characters. Our approach involves processing both facial and bodily features within a unified network architecture, facilitating the extraction of identity-aligned character embeddings that capture individual identities while preserving the effectiveness of face and body features. This integrated character representation enhances feature extraction and improves character re-identification compared to re-identification by face or body independently, offering a parameter-efficient solution. By extensively validating our method using in-series and inter-series evaluation metrics, we demonstrate its effectiveness in consistently re-identifying comic characters. Compared to existing methods, our approach not only addresses the challenge of character re-identification but also serves as a foundation for downstream tasks since it can produce character embeddings without restrictions of face and body availability, enriching the comprehension of comic books. In our experiments, we leverage two newly curated datasets: the 'Comic Character Instances Dataset', comprising over a million character instances and the 'Comic Sequence Identity Dataset', containing annotations of identities within more than 3000 sets of four consecutive comic panels that we collected.
\end{abstract}

\section{Introduction}

Comic character re-identification verifies and links characters across different panels or scenes in a comic book or strip. This task entails accurately connecting instances of a character, even when their appearance varies in terms of pose, expression, or context. Sometimes, characters may be portrayed in close-up shots, focusing predominantly on their faces. In contrast, their faces might not be visible in other instances due to pose variations. Hence, character re-identification in comics is a challenging area that calls for thorough exploration and novel solutions. Despite the significant research in character recognition, the task of character re-identification across comic panels still needs to be explored. Existing literature primarily concentrates on character recognition, particularly emphasizing face recognition and retrieval tasks.

In the context of comic character identification, the primary challenge is the scarcity of accurately annotated data for character faces and bodies, particularly when compared to the available data for character detection \cite{inoue2018cross_comic2k,zheng2020cartoon_icartoon_face,dcm772dataset_digital_comic_indexing}. Assigning identities to comic book characters is challenging due to their arbitrary appearances across pages, resulting in few datasets like Manga109 \cite{manga_109} having character identity annotations.

To address this challenge, we employ a multi-step approach. We begin by using a state-of-the-art model to detect character instances within comic pages. Subsequently, we develop an identity-aware self-supervised model using contrastive learning. This model generates unified embeddings for comic faces and bodies, ensuring consistent identity representation for these character instances. Building on this, we further refine the framework by fine-tuning a linear projector. This projector uses identity annotations collected from sequences of four panels and incorporates metric learning principles and loss functions to extract identity representations. Finally, we cluster identity representations for a given number of characters, where each cluster represents a distinct identity. The core contributions of this work are as follows:

\begin{itemize}
    \item We present a novel \textit{Identity-Aware} self-supervised framework designed to simultaneously process both facial and bodily features of comic book characters resulting in unified character embeddings. This approach exploits the inherent similarity between facial and bodily representations, enabling alignment between the two with \textit{identity-awareness loss}. This approach contributes significantly to the success of solving the character re-identification problem.
    \item By using semi-supervised learning, we fine-tuned a linear layer on top of our identity-aware self-supervised backbone with metric learning techniques, losses and miners, to produce identity features from character embeddings. By clustering identity features, we address the task of comic character re-identification, demonstrating the potential to achieve robust results even with a restricted amount of annotated data on inter-series and, for the first time, in-series evaluation metrics.
    \item We created two new datasets. The first one is the \textit{Comic Character Instances Dataset} containing over a million character instances, which we generated by applying the SOTA \textit{DASS Detector} \cite{topal2022domain_dass} framework to the COMICS \cite{iyyer2017amazing} dataset. The second dataset, named \textit{Comic Sequence Identity Dataset}, includes annotations of identities within more than 3000 sets of four consecutive comic panels that we collected.
\end{itemize}

\section{Related Work}


\subsection{Character Recognition}

The "Progressive Deep Feature Learning" framework proposed by \citeauthor{qin2019progressive} emphasizes the acquisition of a coarse feature representation from unlabeled data, followed by fine-tuning with labeled data. Notably, their approach demonstrates effective performance in manga character recognition, showcasing the potential of leveraging unlabeled data.

The intricate task of learning from diverse modalities is tackled by \citeauthor{zheng2020learning}, who present a meta-continual learning approach for joint face recognition across various styles including sketches, cartoons, caricatures, and real-life photographs. Their shared feature representation and subsequent type-specific classifiers exhibit strong recognition capabilities even in scenarios with limited training data.

\subsection{Character Labeling in Animation}

In the realm of character labeling within animation, \cite{nir2022cast} introduce a self-supervised approach named CAST (Character Labeling in Animation using Self-supervision by Tracking). CAST leverages motion tracking to label characters within animations. This innovative pipeline encompasses automatic shot segmentation, detection, multi-object tracking, and contrastive-based refinement, ultimately defining unique semantic embeddings for animation styles. This approach enhances the understanding of animated content through robust semantic representations.

\subsection{Unsupervised Manga Character Re-identification}

An analogous work to our study is the work by \citeauthor{zhang2022unsupervised_manga_reid}, which investigates unsupervised domain adaptation methods for character re-identification using the Manga109 dataset \cite{manga_109}. Notably, the presence of character identities in Manga109 aids their approach, assisting in image selection and training. In a similar vein, this paper proposes an unsupervised method for manga character re-identification. By extracting face and body features and employing clustering algorithms, it successfully achieves promising re-identification performance without the need for labeled data.

\section{Methodology}

\begin{figure}[t]
    \centering
    \includegraphics[width=0.9\columnwidth]{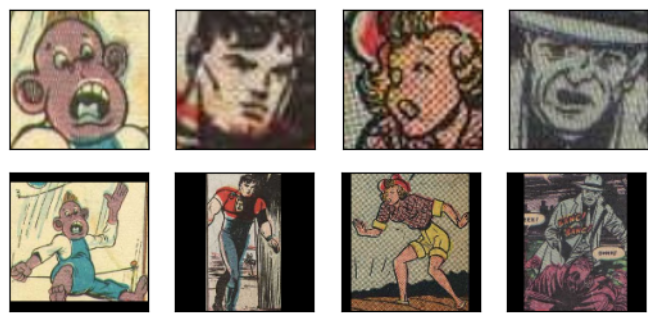}
    \caption{Face and body pair samples from the character instances dataset used in the self-supervision.}
    \label{fig:ssl_dataset_samples}
\end{figure}

\begin{figure*}[t]
    \centering
    \includegraphics[width=0.8\textwidth]{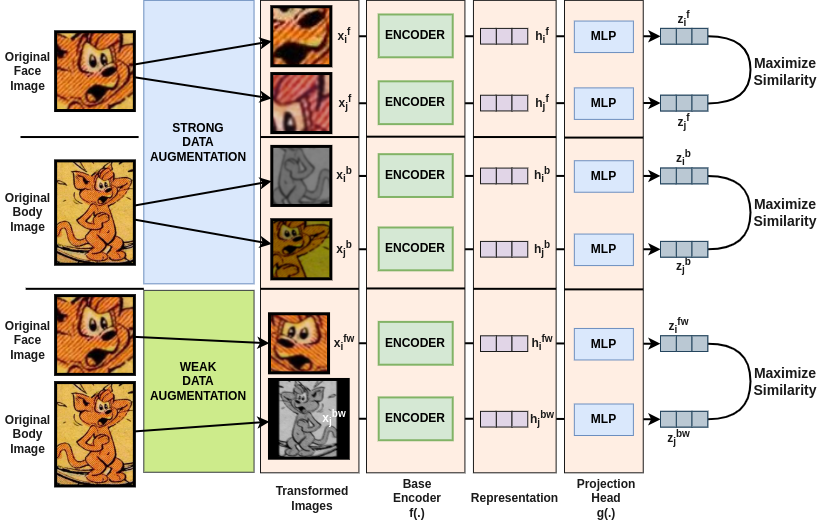}
    \caption{Overview of the Identity-Aware Self-Supervision of Comic Characters approach. The same network architecture is utilized for three distinct tasks, each employing a contrastive learning loss. From top to bottom: (1, 2) Contrastive training of comic face and body images with strong augmentations. (3) Contrastive training of face and body pair images with weak augmentations, ensuring that the embeddings of face and body images depicting the same character exhibit similarity. This encourages both embeddings to encode identity-related information.}
    \label{fig:ssl_id_aware}
\end{figure*}

We created the \textit{COMICS Character Instances} dataset by pairing face, body images using the DASS model from COMICS \cite{iyyer2017amazing}. The \textit{Identity-Aware SimCLR} framework was introduced and trained on this dataset, employing a shared network architecture to simultaneously process face and body images, generating unified and identity-aligned character embeddings. A web tool facilitated partial identity labeling in comic sequences, yielding the "Comic Sequence Identity Dataset." This dataset, in combination with metric learning and contrastive techniques, was then used during semi-supervision to enhance comic character re-identification performance.

\subsection{Character Instances Dataset}

Our primary data source is the COMICS dataset's \cite{iyyer2017amazing} panels. We applied the \textit{XL Stage-3 w/ Mix of Datasets} model from DASS \cite{topal2022domain_dass} to these panels. This model choice was based on its strong performance on both face and body detection tasks on the DCM772 \cite{dcm772dataset_digital_comic_indexing} dataset, which is closely related to COMICS.

The DASS model produces face and body bounding boxes along with detection scores. Subsequent post-processing eliminates low-quality detections. Using these bounding boxes, face and body images were cropped and automatically labeled as character instances, either individually or in pairs, depending on face-body intersections, taking into account overlaps and spatial alignment. The resultant dataset is named \textit{COMICS Character Instances}, with samples showcased in Figure \ref{fig:ssl_dataset_samples}. The \textit{Face+Body 100k} dataset (see \ref{table:ssl_training_counts}) combines 100,000 face and 100,000 body instances as pairs, resulting in 173,828 combined face and body images due to varying face-body availability. Please refer to Appendix B for more details on the dataset.

\begin{table}[t]
    \centering
    \begin{tabular}{@{}ll@{}}
        \hline
        \textbf{Data Type}        & \textbf{Count} \\ \hline
        Face                      & 917,092        \\
        Body                      & 1,139,956      \\
        Face+Body                 & 1,265,710      \\ \hline
        \begin{tabular}[c]{@{}c@{}}Face+Body\\ 100k\end{tabular} & 173,828        \\ \hline
    \end{tabular}
    \caption{Number of images for the character instances dataset. The term '100k' refers to using randomly selected 100,000 faces and bodies. Images of 100k dataset used in the self-supervised pretraining phase. \textit{Face+Body} means face and body pairs, namely character instances.}
    \label{table:ssl_training_counts}
\end{table}

\subsection{Comic Sequence Identity Dataset}

\begin{table}[t]
    \centering
    \begin{tabular}{ccccccc}
        \hline
        \textbf{Sequence} & \textbf{Instance} & \textbf{Face} & \textbf{Body} & \textbf{Identity} \\
        \hline
        3169              & 13716             & 10919         & 13359         & 6720              \\
        \hline
    \end{tabular}
    \caption{Statistics of the Comic Sequence Identity dataset, used for fine-tuning with metric learning for character re-identification tasks.}
    \label{table:seq_id_dataset_counts}
\end{table}

The \textit{Comic Sequence Identity Annotator} is a web-based interface designed to collect identity labels from character instances with diverse facial and bodily resemblances or differences within a sequence of four panels. By clustering characters sharing the same identity, this tool capitalizes on panels with similar contextual cues to obtain partial identity annotations without necessitating full character labeling across the entire comic book. This annotation process produces the \textit{Comic Sequence Identity Dataset}, striking a balance between easy-to-annotate character identities and offering a valuable resource for enhancing machine learning in character-related tasks. This dataset showcases over two unique characters per sequence on average (see Table \ref{table:seq_id_dataset_counts}), reinforcing its suitability for metric learning, since it exploits similarity or dissimilarity between comic character identities effectively. Furthermore,  we expanded the available similar-dissimilar pairs by linking characters across sequences through unique character instance IDs, enabling inter-sequence identity transfer when panels intersect between different sequences. Please refer to Appendix C for more details on the annotator and the dataset.

\subsection{Identity-Aware Self-Supervision of Comic Characters}

In literature, comic character recognition is usually done by the characters' faces. The face-body combination module uses face and body information in Unsupervised Manga Character Re-identification \cite{zhang2022unsupervised_manga_reid}. However, they train separate networks for each part and use each part's information to enhance their soft labels by comparing the probabilities of classes coming from face and body modules. Approaches like this lack the properties of combining features coming from the face and body to create \textit{unified character embeddings}.

To solve those issues, inspired by the contrastive learning approaches, we propose \textbf{Identity-Aware Comic Character Self-supervision} framework, which is a generic framework that can be applied with contrastive learning frameworks. For our experiments we employ and extend SimCLR framework \cite{chen2020simple_simclr_v1,chen2020big_simclr_v2}, thus call it  \textbf{Identity-Aware SimCLR}.

\subsubsection{Identity-Aware SimCLR}

The Identity-Aware SimCLR approach uses a shared network architecture to perform three tasks that aim to capture visual similarities within specific image parts, face-to-face and body-to-body, and encode identity-related information, face-to-body and vice-versa.

The first task involves contrastive training of comic face images with strong augmentations, following the augmentation strategy employed in SimCLR. The network learns to produce representations of comic book faces by maximizing the similarity between augmented versions of the same face image and minimizing the similarity between different faces.

The second task does the same for body images. However, the augmentations are the same except for the image dimensions for resizing. We use 96x96 images for the face, and for the body, it is 128x128. Larger images are used for the body to adequately preserve details for improved representation.

The third task introduces pairs of face and body images depicting the same character. These images undergo weak transformations to ensure that the embeddings of both coarse face and body images exhibit similarity in the feature space. This encourages the network to encode identity-related information in face and body representations. Please refer to Appendix E for details of both augmentations and training details.

By jointly optimizing these tasks using the NT-Xent loss, the Identity-Aware SimCLR approach enables the network to learn representations that capture visual similarities within face and body parts while incorporating identity-related information across them.

In other words, the Identity-Aware SimCLR approach learns to identify visual features that are unique to faces and bodies, as well as features that are shared between faces and bodies. This allows the network to distinguish between different characters, even if they are only shown in part. The loss function for Identity-Aware SimCLR can be seen in Equation \ref{eq:id_aware_nt_xent}.

\begin{equation}
    \mathcal{L}{ij}^{f} = -\log\left(\frac{{\exp(\text{{sim}}(z_{i}^{f}, z_{j}^{f}) / \tau)}}{{\sum_{k=1}^{2N} \mathbb{1}{[k \neq i]} \exp(\text{{sim}}(z_{i}^{f}, z_{k}^{f}) / \tau)}}\right)
\end{equation}

\begin{equation}
    \mathcal{L}{ij}^{b} = -\log\left(\frac{{\exp(\text{{sim}}(z_{i}^{b}, z_{j}^{b}) / \tau)}}{{\sum_{k=1}^{2N} \mathbb{1}{[k \neq i]} \exp(\text{{sim}}(z_{i}^{b}, z_{k}^{b}) / \tau)}}\right)
\end{equation}

\begin{equation}
    \mathcal{L}{ij}^{id} = -\log\left(\frac{{\exp(\text{{sim}}(z_{i}^{f_w}, z_{j}^{b_w}) / \tau)}}{{\sum_{k=1}^{2N} \mathbb{1}{[k \neq i]} \exp(\text{{sim}}(z_{i}^{f_w}, z_{k}^{b_w}) / \tau)}}\right)
\end{equation}

where \(z\) values are as follows. \(z^{f}\): Projected features of face with strong augmentations, \(z^{b}\): Projected features of body with strong augmentations, \(z^{f_w}\) or \(z^{b_w}\): Projected features of face or body with weak augmentations. The overall loss function for identity-aware SimCLR is given by the summation of face, body and \textit{identity-awareness} losses:

\begin{equation}\label{eq:id_aware_nt_xent}
    \mathcal{L}{ij} = \mathcal{L}{ij}^{f} + \mathcal{L}{ij}^{b} + \mathcal{L}{ij}^{id}
\end{equation}

We use ResNet-50 \cite{he2016deep_resnet} as the base encoder, which extracts representation vectors from augmented data examples. Additionally, we utilize a deeper MLP projection head, following the architecture suggested in SimCLR v2 \cite{chen2020big_simclr_v2}. This projection head, denoted as $g(\cdot)$, maps these representations to a space where the contrastive loss is applied.


\subsection{Semi-Supervision with Metric Learning for Re-Identification}

Metric learning is a machine learning approach that focuses on learning a distance or similarity metric between data points in a meaningful way for a specific task. The primary goal of metric learning is to ensure that similar data points are closer to each other in the learned metric space, while dissimilar data points are farther apart \cite{kaya2019deep_metric_survey}. This learned metric can be used for various applications, such as classification, clustering, retrieval, and re-identification tasks. In our context, similarity refers to belonging to the same character identity, while dissimilarity means having a different character identity.

To address the comic character re-identification task, we adopted the subsequent strategy. After self-supervised model training, we retained the core encoder $f(\cdot)$ and the projection head until the middle layer—discarding the rest, denoted as $f'(\cdot)$ which is frozen during fine-tuning. Extending the network with a linear layer, we applied L2 normalization for identity representations, akin to FaceNet's pioneering approach \cite{schroff2015facenet}, designated as $g_{\text{id}}(\cdot)$. Subsequently, we formulated a coherent character feature extraction process as follows: Let $e_{\text{face}} = f'(x_{\text{face}})$ represent face features, and $e_{\text{body}} = f'(x_{\text{body}})$ embody body features for a character. If any of the futures is absent, a zero vector serves as a replacement. For face-body feature fusion, methodologies like vector summation or concatenation can be applied. Notably, our experiments demonstrate that summation yields better results, likely due to the identity-awareness loss bridging and aligning face and body features, enhancing identity encoding. Consequently, the ultimate identity embedding is obtained through $e_{\text{id}} = g_{\text{id}}(e_{\text{face}} + e_{\text{body}})$.

For semi-supervision, we employed metric learning methods and leveraged contrastive approaches, which are suitable for our data collection methodology. When constructing our metric learning pipeline, we benefited from the PyTorch Metric Learning library \cite{Musgrave2020PyTorchML_pml}. We trained our model using well-suited loss functions. Notably, the Contrastive loss \cite{hadsell2006dimensionality_contrastive_loss}, the Triplet-Margin loss \cite{balntas2016learning_triplet_margin_loss}, Intra-Pair Variance loss \cite{yu2019deep_tuplet_margin_loss}, Multi-Similarity loss \cite{wang2019multi_similarity_miner}. A critical aspect of metric learning involves determining an appropriate mining strategy for selecting similar or dissimilar items that contribute to the loss function. This strategy significantly influences the success of the metric learning process. Multi-Similarity miner \cite{wang2019multi_similarity_miner} is selected as the base miner, it is used in conjunction with Triplet-Margin and Multi-Similarity losses. Please refer to Appendix F for more details.

\subsubsection{Meta-Mining Strategy for Comic Sequences}

We introduced a \textit{meta-mining} strategy to improve our metric learning pipeline. So that the adverse effects of using comparative identity labels are mitigated. This approach establishes character relationships using cliques formed from all character identities and ensures the accuracy of mining similar and dissimilar pairs. The outcome of meta-mining is the indices in minibatch that would be mined together. Then the base mining algorithm is applied. Without the meta-mining process, there is a risk of using character instances as 'dissimilar' pairs since our labels cannot guarantee identities in different sequences. In conjunction with meta-mining, the \textit{mix-series} approach involves adding character instances from series that are not involved in the cliques as dissimilar instances allowing for the selection of pairs that leads to better generalization. Please refer to Appendix D for more details on the meta-mining algorithm.

\subsection{Identity Assignment Across Panels with Character Clustering for Re-Identification}


In a sequence of four panels, character face and body images are paired if possible to create character instances and then processed through the semi-supervised model, producing identity embeddings ($e_{\text{id}}$) for each. These embeddings are then utilized by a clustering algorithm to determine cluster membership, indicating distinct character identities.

The challenge lies in selecting an appropriate clustering approach, as traditional methods like K-means require a predefined number of clusters which is unsuitable due to the varying number of character identities in a sequence. Thus, we explored options such as DBSCAN \cite{ester1996density_dbscan} and Agglomerative Clustering, ultimately adopting Agglomerative Clustering \footnote{implemented using scikit-learn \cite{scikit-learn}} due to its compatibility with the sample size and characteristics of our problem. Please refer to Appendix A for qualitative evaluation and Appendix H for clustering details.

\section{Experimental Setup}

Our experimental design comprises three primary setups. The initial configuration aims to contrast our method with other baseline approaches, involving a fine-tuned ResNet50 \cite{he2016deep_resnet} backbone with a linear identity layer, extraction of color histogram vectors, and a random baseline. This comparison spans both our self-supervised and semi-supervised models. In the case of the self-supervised model, no additional training was conducted; instead, we utilized the summation of projection head outputs, $g_{\text{id}}(\cdot)$. If there was training, Triplet-Margin loss was used with a Multi-similarity miner.


The second experimental setup can be interpreted as ablation studies conducted from various perspectives, including experiments to comprehend the impact of identity-awareness. It also seeks to address questions like whether character identity is conveyed more dominantly through the face or the body in comics. To achieve this, we trained a set of models within the same framework by altering only the data and loss functions, resulting in distinct outcomes. These models include the following: the \textbf{Face-Only} and \textbf{Body-Only} models, trained with either face or body data during self-supervision and semi-supervision processes; the \textbf{Unaligned} model, where both face and body data are involved in training but the identity-awareness loss is excluded from self-supervision; the \textbf{Aligned} model, which fully employs both face and body data with added identity-awareness loss; and finally, the \textbf{Separate} model, which employs separate self-supervision processes for face and body backbones while integrating their embeddings through fine-tuning. The Separate model encompasses two backbone models, doubling the parameter count. The efficiency of the Aligned model in terms of parameters eliminates this need, yielding superior results compared to the Unaligned model.

The final setup comprises experiments conducted to observe the effects of loss and miner in the metric learning and fine-tuning processes.

\subsection{Evaluation Strategies}

We propose two distinct evaluation strategies to assess the performance of our models: local (in-series) and global (inter-series) approaches. These strategies offer different perspectives when evaluating the effectiveness of our models. The common approach in the literature is to evaluate the performance of the model globally, and there is no evaluation in the literature like the proposed local method. It directly indicates the re-identification performance across panels. The success of this is directly proportional to the success of tracking with character identification within a series and sequential panels.

\begin{itemize}
    \item \textbf{Local (In-Series)}: This approach considers character evaluation within a sequence of four consecutive panels, having an average of four character instances from two distinct identities (see Table \ref{table:seq_id_dataset_counts}). We adopt this strategy to showcase our models' performance in character re-identification across sequential panels. Local evaluation is essential as \textbf{it reflects the spatiotemporal character re-identification performance}.
    \item \textbf{Global (Inter-Series)}: In this approach, we evaluate the discriminative power of our models in a broader context. Instead of assessing accuracies within sequences, we select query and reference character instances from different comic series in the test set. The objective is to retrieve reference instances based on similarity measures such as L2 or cosine similarity. The dataset consists of \textbf{126 query instances} and \textbf{361 reference instances}. For a query character instance, there are on average ~1.5 references.
\end{itemize}

The following evaluation metrics frequently utilized in information retrieval, which effectively characterize the embedding space \cite{musgrave2020metric, Musgrave2020PyTorchML_pml}, are employed to measure the performance of our models in conjunction with our evaluation strategies: Mean Average Precision (MAP), Mean Average Precision at R (MAP@R), Mean Reciprocal Rank (MRR), Precision@1 (P@1), R-precision (R-P).

\begin{table*}[t]
    \centering
    \begin{tabular}{@{}ccccccccccccc@{}}
        \toprule
        \textbf{Method}                           &
        \multicolumn{3}{c}{\textbf{NT-Xent Loss}} &
        \multicolumn{3}{c}{\textbf{Top-1 (\%)}}   &
        \multicolumn{3}{c}{\textbf{Top-5 (\%)}}   &
        \multicolumn{3}{c}{\textbf{Mean Pos.}}      \\ \cmidrule(l){2-13}
                                                  &
        Face                                      &
        Body                                      &
        Identity                                  &
        Face                                      &
        Body                                      &
        Identity                                  &
        Face                                      &
        Body                                      &
        Identity                                  &
        Face                                      &
        Body                                      &
        Identity                                    \\ \midrule
        FACE                                      &
        0.198                                     &
        -                                         &
        -                                         &
        97.0                                      &
        -                                         &
        -                                         &
        99.3                                      &
        -                                         &
        -                                         &
        1.2                                       &
        -                                         &
        -                                           \\
        BODY                                      &
        -                                         &
        0.409                                     &
        -                                         &
        -                                         &
        93.8                                      &
        -                                         &
        -                                         &
        97.9                                      &
        -                                         &
        -                                         &
        1.6                                       &
        -                                           \\
        UNALIGNED                                 &
        0.194                                     &
        0.398                                     &
        6.540                                     &
        \textbf{97.7}                             &
        94.5                                      &
        22.4                                      &
        \textbf{99.5}                             &
        98.3                                      &
        31.3                                      &
        \textbf{1.1}                              &
        1.6                                       &
        147.9                                       \\
        ALIGNED                                   &
        \textbf{0.192}                            &
        \textbf{0.368}                            &
        \textbf{0.207}                            &
        97.5                                      &
        \textbf{94.9}                             &
        \textbf{97.3}                             &
        99.4                                      &
        \textbf{98.6}                             &
        \textbf{99.5}                             &
        1.1                                       &
        \textbf{1.5}                              &
        \textbf{1.1}                                \\ \bottomrule
    \end{tabular}
    \caption{Results of self-supervised training experiments conducted on face, body, and combining face and body for comic characters. The table presents loss values, top-k accuracy metrics, and mean position for each task, self-supervision of face, body, and identity. The models serve as the backbone for character re-identification tasks. The difference between aligned and unaligned methods is the existence of Identity-Awareness loss during training.}
    \label{table:ssl_results}
\end{table*}

\section{Results and Discussion}

\subsection{Self-Supervised Learning of Comic Characters}

We evaluate self-supervised models trained on comic character images by analyzing loss values, top-k accuracy metrics, and mean position determined through cosine similarity within the batch. Although the key assessment of self-supervision rests on their performance in downstream tasks, the selected loss function, aiming to bring closer projection features of transformed images from the same source, aligns well with the metric learning objective of character re-identification. Hence, the results (refer to Table \ref{table:ssl_results}) offer insights into model efficacy.

From the facial results, models trained with both face and body data exhibit enhanced performance in loss and top-k accuracy, indicating the beneficial role of body data in training for improving face representations. Similar patterns are observed in body performance comparisons. Incorporating face data alongside body data during training elevates body representations. Even the \textit{Unaligned} model, with considerably lower identity performance compared to the aligned version, still achieves a noteworthy 22.4\% top-1 accuracy within a batch of 320 samples.

To assess the effects of \textit{Identity-Alignment}, it is best to compare the performance of the models, Aligned vs Unaligned, in terms of their ability to capture identity-related information. The identity-aware module is far superior to its unaligned counterpart.  This distinction arises solely from the \textit{Identity-Awareness Loss}. It can be argued that the relatively better performance of the unaligned models, even without the identity-awareness loss, creates a favorable environment for the success of identity alignment of faces and bodies.

Unaligned models marginally outperform aligned ones in face performance in top-k, while aligned models excel in all metrics for body data. This hints that identity awareness predominantly benefits distinguishing body features in self-supervision performance. Notably, the consistently superior performance of the identity-aligned model across all evaluation metrics for identity results reinforces its potential in character re-identification tasks. Since it shows the model's representational ability to retrieve given a face or body its counterpart.

\subsection{Semi-Supervision with Metric Learning }

\begin{table}[t]
    \centering
    \begin{tabular}{lccccc}
        \toprule
        \multicolumn{1}{c}{\textbf{}}       & \textbf{map}                        & \textbf{map@r} & \textbf{mrr}  & \textbf{p@1}  & \textbf{r-p}  \\ \cmidrule(l){2-6}
        \multicolumn{1}{c}{\textbf{Method}} & \multicolumn{5}{c}{\textbf{Local}}                                                                   \\
        \midrule
        Random                              & 57.5                                & 30.6           & 61.7          & 35.6          & 38.8          \\
        Color Hist.                         & 58.2                                & 32.7           & 62.8          & 40.9          & 38.9          \\
        ResNet50                            & 70.4                                & 49.7           & 74.6          & 58.8          & 54.0          \\
        Self-S. (ours)                      & 66.8                                & 46.2           & 70.3          & 49.8          & 53.0          \\
        Semi-S. (ours)                      & \textbf{78.7}                       & \textbf{60.9}  & \textbf{81.5} & \textbf{68.1} & \textbf{64.7} \\
        \midrule
                                            & \multicolumn{5}{c}{\textbf{Global}}                                                                  \\
        \midrule
        Random                              & 2.2                                 & 0.2            & 2.8           & 0.4           & 0.3           \\
        Color Hist.                         & 5.8                                 & 2.9            & 8.0           & 3.9           & 3.3           \\
        ResNet50                            & 21.2                                & 13.8           & 25.1          & 15.8          & 14.9          \\
        Self-S. (ours)                      & \textbf{66.8}                       & \textbf{46.2}  & \textbf{70.3} & \textbf{49.8} & \textbf{53.0} \\
        Semi-S. (ours)                      & 42.5                                & 31.6           & 47.4          & 35.7          & 33.2          \\
        \bottomrule
    \end{tabular}
    \caption{Baseline models' results in terms of performance metrics for the local and global evaluation.}
    \label{table:pml_baseline_results}
\end{table}

\subsubsection{Baselines}

Table \ref{table:pml_baseline_results} compiles baseline model performance, assessed both on a global and local scale. The Random model's minimal performance results from a surplus of reference instances per query, leading to subdued global evaluation metrics. In local evaluation, P@1 confirms the average presence of two character pairs per sequence.

The semi-supervised model excels locally among baselines, most importantly it is superior to the supervised ResNet50 model with a linear layer, its direct competitor since it is the backbone for it. Though it balances local and global accuracy trade-offs when observing the global performance of the self-supervised setup. Consequently, our approach outperforms basics and underscores the importance of balancing spatiotemporal and universal aspects of comic character re-identification during fine-tuning.

\subsubsection{Effects of Face, Body, and Identity-Awareness on Character Re-Identification}

\begin{figure*}[t]
    \centering
    \includegraphics[width=0.9\textwidth]{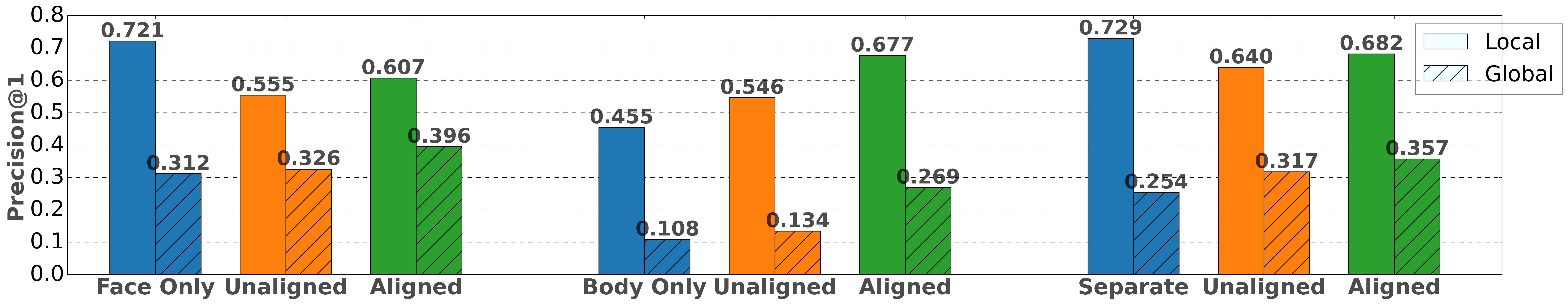}
    \caption{Linear evaluation of representations of different self-supervised backbones on the face, body, and unified face-body data from left to right.}
    \label{fig:pml_all_performance}
\end{figure*}

Figure \ref{fig:pml_all_performance} shows the performance of face, body, and unified face-body representations for precision@1 metric. The same trends are consistent with all other metrics we use. However, precision@1 is selected for demonstration since it is the most important metric of all, deciding the performance for character re-identification. For face re-identification, in local metrics, the face-only model consistently outperforms other models, while globally superior performance is observed in the aligned model trained with face and body data. Surprisingly, the unaligned model performs better globally than the face-only model. The identity-awareness loss might enhance discriminative features related to character identities, with the aligned model benefiting from contextual cues relevant to the entire character. The face-only model excels in extracting spatial and temporal context, contributing to its strong local performance. The aligned module's improvement over the unaligned for both evaluation strategies validates our method, showing identity-awareness loss improves re-identification and recognition of faces.

Contrasting the face evaluation, the evaluation of models based solely on body data shows consistent rankings in both local and global assessments. The aligned module outperforms the unaligned and body-only models, indicating that extracting identity information from the body alone is more complex compared to the face. However, the face's prominence in carrying identity information leads to improved performance even when only body data is used, attributed to the model's training with both faces and bodies.

The trend observed in the face-only evaluation remains consistent when evaluating face and body data. Separately trained self-supervised learning backbones with fine-tuned linear layers for face and body data exhibit better local performance than simultaneous training of both with the cost of doubling the number of parameters. The aligned module consistently outperforms the unaligned counterpart in both local and global evaluations and performs best globally, suggesting the effectiveness of our method. Notably, there is a positive difference compared to the face-only evaluation. The aligned and separate models exhibit similar closer local results, suggesting efficient distribution of spatiotemporal information across the face and body. Utilizing both sources of information allows the aligned model to maximize its potential in local and global contexts.

\subsubsection{Losses and Mining for Metric Learning}

\begin{table}[t]
    \centering
    \begin{tabular}{lccccc}
        \toprule
        \multicolumn{1}{c}{\textbf{}}     & \textbf{map}                        & \textbf{map@r} & \textbf{mrr}  & \textbf{p@1}  & \textbf{r-p}  \\ \cmidrule(l){2-6}
        \multicolumn{1}{c}{\textbf{Loss}} & \multicolumn{5}{c}{\textbf{Local}}                                                                   \\ \midrule
        Contrastive                       & 77.7                                & 58.4           & 80.4          & 65.3          & 62.7          \\
        Intra-Pair Var.                   & 74.0                                & 53.5           & 76.5          & 59.1          & 58.6          \\
        Triplet-Margin                    & \textbf{79.6}                       & \textbf{62.5}  & 81.6          & 68.1          & \textbf{66.0} \\
        \hspace{0.2cm}+mix-series         & 79.0                                & 61.4           & \textbf{81.7} & \textbf{68.4} & 65.1          \\
        Multi-Sim.                        & 78.4                                & 59.7           & 81.1          & 67.1          & 63.8          \\
        \hspace{0.2cm}+mix-series         & 78.9                                & 59.7           & 81.2          & 67.0          & 63.5          \\ \midrule
                                          & \multicolumn{5}{c}{\textbf{Global}}                                                                  \\ \midrule
        Contrastive                       & \textbf{64.6}                       & \textbf{51.7}  & \textbf{69.9} & \textbf{58.4} & \textbf{52.9} \\
        Intra-Pair Var.                   & 59.7                                & 45.0           & 64.6          & 51.7          & 46.3          \\
        Triplet-Margin                    & 40.0                                & 28.9           & 46.2          & 35.0          & 30.1          \\
        \hspace{0.2cm}+mix-series         & 61.9                                & 48.3           & 66.2          & 54.4          & 49.3          \\
        Multi-Sim.                        & 47.9                                & 36.2           & 53.0          & 41.1          & 37.6          \\
        \hspace{0.2cm}+mix-series         & 59.1                                & 45.2           & 62.8          & 50.0          & 46.4          \\ \bottomrule
    \end{tabular}
    \caption{Linear evaluation trained with different metric learning loss functions.}
    \label{table:pml_loss_miner_results}
\end{table}

Table \ref{table:pml_loss_miner_results} shows the results of fine-tuning experiments with different loss functions and mining strategies. Comparing loss functions with and without miners, we found that Triplet-Margin and Multi-Similarity losses, which use miners, improve local performance but worsen global performance. The "mix-series" augmentation to mining balances this since it improves global performance. When no miner is used, any character can be paired, which can lead to wrong pairs due to data collection. This might be why miner-based losses show better local performance. Contrastive loss works globally and competently locally, capturing relationships between face and body images. Intra-pair variance loss is similar but slightly worse. It might not handle noise from incorrect labels as well as contrastive loss. Please refer to Appendix G for more details on fine-tuning decisions.

\section{Conclusion \& Future Work}

In this work, we introduce a pioneering \textit{Identity-Aware} parameter efficient self-supervised framework for character re-identification in comics. By processing facial and bodily features together, our approach unifies character embeddings while aligning these representations through identity-awareness. Leveraging semi-supervised and metric learning techniques, we demonstrate its effectiveness in consistently re-identifying characters, even with limited annotated data. Our contributions extend to the creation of two new datasets, encompassing character instances and identity annotations within sequential panels. This framework not only enhances character re-identification but also offers potential for downstream applications in comic analysis and beyond.

Future exploration could involve utilizing character and identity representations for diverse tasks beyond character re-identification. Integrating character and identity knowledge into scene understanding tasks could enhance character relationship and interaction comprehension. Furthermore, character embeddings could be harnessed for generating comic text and images, resulting in more coherent and contextually harmonious narratives. These avenues hold the potential to advance comic comprehension, generation, and interactivity, offering novel directions for future research.

\bibliography{aaai24}

\begin{thebibliography}{24}
\providecommand{\natexlab}[1]{#1}

\bibitem[{Aizawa et~al.(2020)Aizawa, Fujimoto, Otsubo, Ogawa, Matsui, Tsubota,
  and Ikuta}]{manga_109}
Aizawa, K.; Fujimoto, A.; Otsubo, A.; Ogawa, T.; Matsui, Y.; Tsubota, K.; and
  Ikuta, H. 2020.
\newblock Building a Manga Dataset ``Manga109'' with Annotations for Multimedia
  Applications.
\newblock \emph{IEEE MultiMedia}, 27(2): 8--18.

\bibitem[{Balntas et~al.(2016)Balntas, Riba, Ponsa, and
  Mikolajczyk}]{balntas2016learning_triplet_margin_loss}
Balntas, V.; Riba, E.; Ponsa, D.; and Mikolajczyk, K. 2016.
\newblock Learning local feature descriptors with triplets and shallow
  convolutional neural networks.
\newblock In \emph{Bmvc}, volume~1, 3.

\bibitem[{Chen et~al.(2020{\natexlab{a}})Chen, Kornblith, Norouzi, and
  Hinton}]{chen2020simple_simclr_v1}
Chen, T.; Kornblith, S.; Norouzi, M.; and Hinton, G. 2020{\natexlab{a}}.
\newblock A Simple Framework for Contrastive Learning of Visual
  Representations.
\newblock \emph{arXiv preprint arXiv:2002.05709}.

\bibitem[{Chen et~al.(2020{\natexlab{b}})Chen, Kornblith, Swersky, Norouzi, and
  Hinton}]{chen2020big_simclr_v2}
Chen, T.; Kornblith, S.; Swersky, K.; Norouzi, M.; and Hinton, G.
  2020{\natexlab{b}}.
\newblock Big Self-Supervised Models are Strong Semi-Supervised Learners.
\newblock \emph{arXiv preprint arXiv:2006.10029}.

\bibitem[{Deng et~al.(2009)Deng, Dong, Socher, Li, Li, and
  Fei-Fei}]{deng2009imagenet}
Deng, J.; Dong, W.; Socher, R.; Li, L.-J.; Li, K.; and Fei-Fei, L. 2009.
\newblock Imagenet: A large-scale hierarchical image database.
\newblock In \emph{2009 IEEE conference on computer vision and pattern
  recognition}, 248--255. Ieee.

\bibitem[{Ester et~al.(1996)Ester, Kriegel, Sander, Xu
  et~al.}]{ester1996density_dbscan}
Ester, M.; Kriegel, H.-P.; Sander, J.; Xu, X.; et~al. 1996.
\newblock A density-based algorithm for discovering clusters in large spatial
  databases with noise.
\newblock In \emph{kdd}, volume~96, 226--231.

\bibitem[{Hadsell, Chopra, and
  LeCun(2006)}]{hadsell2006dimensionality_contrastive_loss}
Hadsell, R.; Chopra, S.; and LeCun, Y. 2006.
\newblock Dimensionality reduction by learning an invariant mapping.
\newblock In \emph{2006 IEEE computer society conference on computer vision and
  pattern recognition (CVPR'06)}, volume~2, 1735--1742. IEEE.

\bibitem[{He et~al.(2016)He, Zhang, Ren, and Sun}]{he2016deep_resnet}
He, K.; Zhang, X.; Ren, S.; and Sun, J. 2016.
\newblock Deep residual learning for image recognition.
\newblock In \emph{Proceedings of the IEEE conference on computer vision and
  pattern recognition}, 770--778.

\bibitem[{Inoue et~al.(2018)Inoue, Furuta, Yamasaki, and
  Aizawa}]{inoue2018cross_comic2k}
Inoue, N.; Furuta, R.; Yamasaki, T.; and Aizawa, K. 2018.
\newblock Cross-domain weakly-supervised object detection through progressive
  domain adaptation.
\newblock In \emph{Proceedings of the IEEE conference on computer vision and
  pattern recognition}, 5001--5009.

\bibitem[{Iyyer et~al.(2017)Iyyer, Manjunatha, Guha, Vyas, Boyd-Graber, au2,
  and Davis}]{iyyer2017amazing}
Iyyer, M.; Manjunatha, V.; Guha, A.; Vyas, Y.; Boyd-Graber, J.; au2, H. D.~I.;
  and Davis, L. 2017.
\newblock The Amazing Mysteries of the Gutter: Drawing Inferences Between
  Panels in Comic Book Narratives.
\newblock arXiv:1611.05118.

\bibitem[{Kaya and Bilge(2019)}]{kaya2019deep_metric_survey}
Kaya, M.; and Bilge, H.~{\c{S}}. 2019.
\newblock Deep metric learning: A survey.
\newblock \emph{Symmetry}, 11(9): 1066.

\bibitem[{Musgrave, Belongie, and Lim(2020{\natexlab{a}})}]{musgrave2020metric}
Musgrave, K.; Belongie, S.; and Lim, S.-N. 2020{\natexlab{a}}.
\newblock A metric learning reality check.
\newblock In \emph{Computer Vision--ECCV 2020: 16th European Conference,
  Glasgow, UK, August 23--28, 2020, Proceedings, Part XXV 16}, 681--699.
  Springer.

\bibitem[{Musgrave, Belongie, and
  Lim(2020{\natexlab{b}})}]{Musgrave2020PyTorchML_pml}
Musgrave, K.; Belongie, S.~J.; and Lim, S.-N. 2020{\natexlab{b}}.
\newblock PyTorch Metric Learning.
\newblock \emph{ArXiv}, abs/2008.09164.

\bibitem[{Nguyen, Rigaud, and
  Burie(2018)}]{dcm772dataset_digital_comic_indexing}
Nguyen, N.-V.; Rigaud, C.; and Burie, J.-C. 2018.
\newblock Digital Comics Image Indexing Based on Deep Learning.
\newblock \emph{Journal of Imaging}, 4(7).

\bibitem[{Nir, Rapoport, and Shamir(2022)}]{nir2022cast}
Nir, O.; Rapoport, G.; and Shamir, A. 2022.
\newblock CAST: Character labeling in Animation using Self-supervision by
  Tracking.
\newblock In \emph{Computer Graphics Forum}, volume~41, 135--145. Wiley Online
  Library.

\bibitem[{Pedregosa et~al.(2011)Pedregosa, Varoquaux, Gramfort, Michel,
  Thirion, Grisel, Blondel, Prettenhofer, Weiss, Dubourg, Vanderplas, Passos,
  Cournapeau, Brucher, Perrot, and Duchesnay}]{scikit-learn}
Pedregosa, F.; Varoquaux, G.; Gramfort, A.; Michel, V.; Thirion, B.; Grisel,
  O.; Blondel, M.; Prettenhofer, P.; Weiss, R.; Dubourg, V.; Vanderplas, J.;
  Passos, A.; Cournapeau, D.; Brucher, M.; Perrot, M.; and Duchesnay, E. 2011.
\newblock Scikit-learn: Machine Learning in {P}ython.
\newblock \emph{Journal of Machine Learning Research}, 12: 2825--2830.

\bibitem[{Qin et~al.(2019)Qin, Zhou, Li, Wang, Wang, and
  Tang}]{qin2019progressive}
Qin, X.; Zhou, Y.; Li, Y.; Wang, S.; Wang, Y.; and Tang, Z. 2019.
\newblock Progressive deep feature learning for manga character recognition via
  unlabeled training data.
\newblock In \emph{Proceedings of the ACM Turing Celebration Conference-China},
  1--6.

\bibitem[{Schroff, Kalenichenko, and Philbin(2015)}]{schroff2015facenet}
Schroff, F.; Kalenichenko, D.; and Philbin, J. 2015.
\newblock Facenet: A unified embedding for face recognition and clustering.
\newblock In \emph{Proceedings of the IEEE conference on computer vision and
  pattern recognition}, 815--823.

\bibitem[{Topal, Yuret, and Sezgin(2022)}]{topal2022domain_dass}
Topal, B.~B.; Yuret, D.; and Sezgin, T.~M. 2022.
\newblock Domain-Adaptive Self-Supervised Pre-Training for Face \& Body
  Detection in Drawings.
\newblock \emph{arXiv preprint arXiv:2211.10641}.

\bibitem[{Wang et~al.(2019)Wang, Han, Huang, Dong, and
  Scott}]{wang2019multi_similarity_miner}
Wang, X.; Han, X.; Huang, W.; Dong, D.; and Scott, M.~R. 2019.
\newblock Multi-similarity loss with general pair weighting for deep metric
  learning.
\newblock In \emph{Proceedings of the IEEE/CVF conference on computer vision
  and pattern recognition}, 5022--5030.

\bibitem[{Yu and Tao(2019)}]{yu2019deep_tuplet_margin_loss}
Yu, B.; and Tao, D. 2019.
\newblock Deep metric learning with tuplet margin loss.
\newblock In \emph{Proceedings of the IEEE/CVF international conference on
  computer vision}, 6490--6499.

\bibitem[{Zhang, Wang, and Hu(2022)}]{zhang2022unsupervised_manga_reid}
Zhang, Z.; Wang, Z.; and Hu, W. 2022.
\newblock Unsupervised Manga Character Re-identification via Face-body and
  Spatial-temporal Associated Clustering.
\newblock \emph{arXiv preprint arXiv:2204.04621}.

\bibitem[{Zheng et~al.(2020{\natexlab{a}})Zheng, Yan, Wang, and
  Gou}]{zheng2020learning}
Zheng, W.; Yan, L.; Wang, F.-Y.; and Gou, C. 2020{\natexlab{a}}.
\newblock Learning from the past: Meta-continual learning with knowledge
  embedding for jointly sketch, cartoon, and caricature face recognition.
\newblock In \emph{Proceedings of the 28th ACM International Conference on
  Multimedia}, 736--743.

\bibitem[{Zheng et~al.(2020{\natexlab{b}})Zheng, Zhao, Ren, Yan, Lu, Liu, and
  Li}]{zheng2020cartoon_icartoon_face}
Zheng, Y.; Zhao, Y.; Ren, M.; Yan, H.; Lu, X.; Liu, J.; and Li, J.
  2020{\natexlab{b}}.
\newblock Cartoon face recognition: A benchmark dataset.
\newblock In \emph{Proceedings of the 28th ACM international conference on
  multimedia}, 2264--2272.

\end{thebibliography}

\newpage

\setcounter{secnumdepth}{2}
\appendix

\section*{Appendix}

\section{Qualitative Evaluation}
\label{section:qualitative_eval}

\begin{figure*}[!htb]
    \centering
    \includegraphics[width=0.9\textwidth]{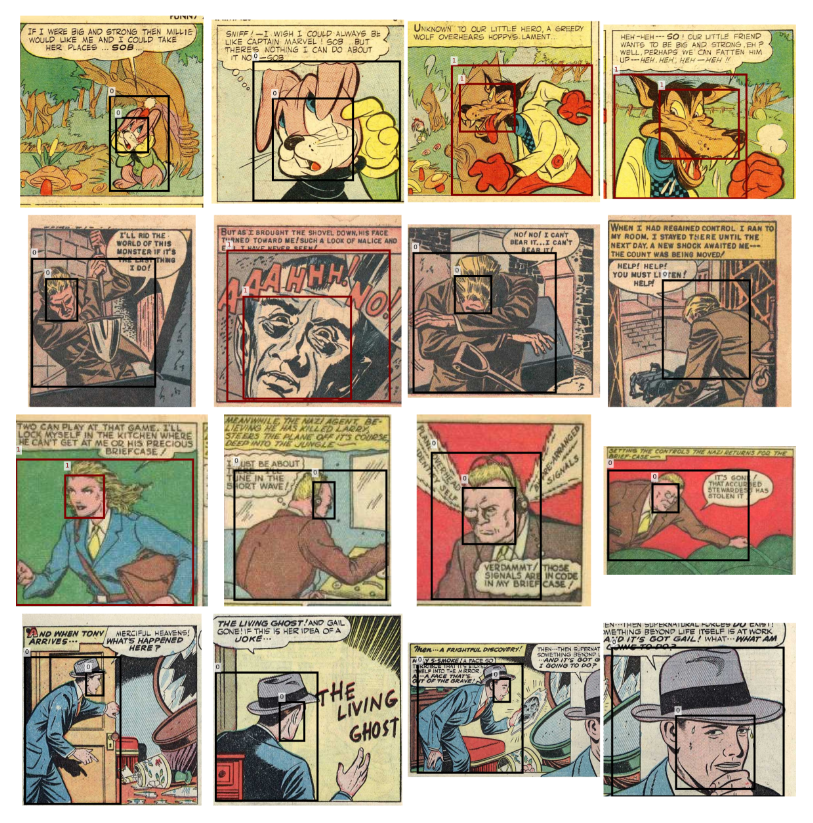}
    \caption{Examples of 4-panel sequences that the model demonstrates successful character reidentification. Each color for a bounding box represents a unique assigned character identity.}
    \label{fig:pml_good_reid}
\end{figure*}

\begin{figure*}[!htb]
    \centering
    \includegraphics[width=0.9\textwidth]{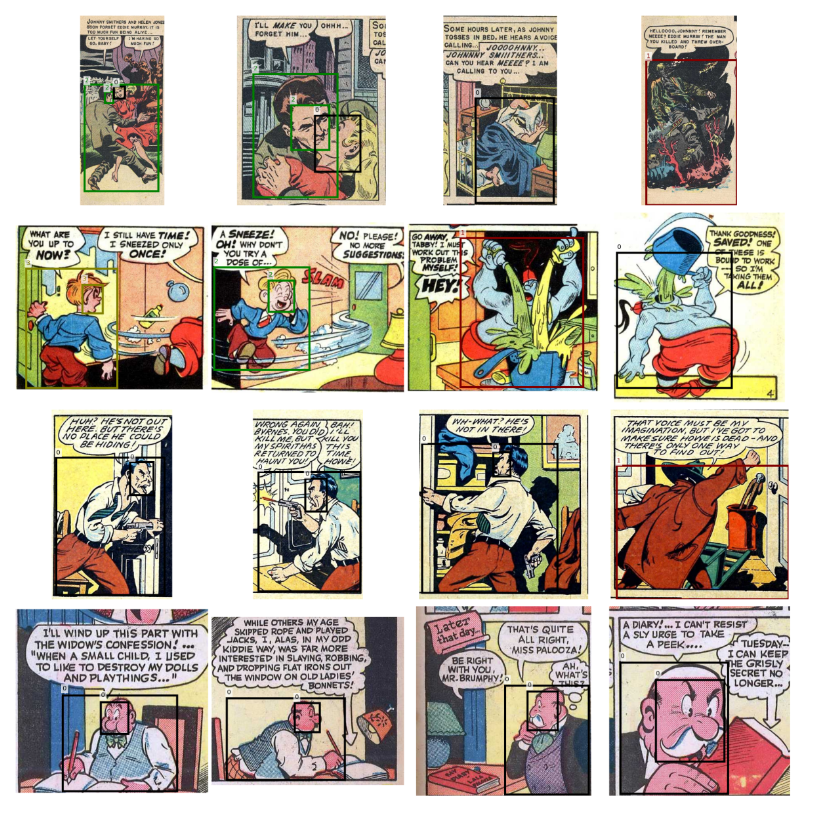}
    \caption{Examples of 4-panel sequences that highlight the challenges faced by the model for character reidentification. Each color for a bounding box represents a unique assigned character identity.}
    \label{fig:pml_bad_reid}
\end{figure*}

Qualitative evaluation is performed on 4-panel sequences. Both successful and problematic examples can be seen in Figures \ref{fig:pml_good_reid} and \ref{fig:pml_bad_reid}. How we assign characters are explained in Section \ref{seciton:identity_assignment}.

Our observations are as follows: In the well-performing examples, characters generally have face and body representations, and their drawing distances are relatively consistent with the scene. Even if they are not the same, there is consistency in facial expressions.

The problematic examples help to understand the challenges faced by the model. However, identifying specific patterns is challenging. The following can be observed if the problematic examples are examined case by case; The numbering is aligned with the order of sequences in Figure \ref{fig:pml_bad_reid}:

\begin{enumerate}
    \item The first two panels in the first sequence indicate that the model performs very well, even separating overlapping characters. However, in the third panel, the male character is assigned to the female identity in the first two panels. My assumption here is that since the female character does not have a body, the model might have made the similarity based on the skin color and complexity and the presence of a body. This might suggest that it is important for both face and body to be detected in the examples.
    \item The example where the model performs the worst in the sequence is when the child character turns around in the first panel and faces forward in the second panel. The model may have difficulty distinguishing complete rotations. For the other character, Negative outcomes can be based on the difficulty of determining identities only based on bodies and having a non-human character.
    \item Here, the mistake made by the model could be the character both turning its back and changing clothes. Otherwise, the model successfully assigns the correct identity to the character instances.
    \item In the last sequence, characters who are quite similar physically and in terms of their faces are presented but are assigned the same identity.
\end{enumerate}

\section{Character Instances Dataset}

The scores and bounding boxes can be derived from the predictions made by the DASS \cite{topal2022domain_dass} model. A series of post-processing steps were undertaken to filter and organize the detected faces and bodies. Inclusion in the self-supervision dataset was decided based on specific criteria: bounding box areas less than 64 and detection scores lower than 0.95 were not considered.

During the training phase, facial images underwent a scaling of 1.2, followed by cropping to a square shape based on the longest side. This was done to ensure complete capture of the character's head. Faces and bodies identified as character instances were automatically labeled. Our approach involved the following steps: If a face intersected with a body in a panel by more than 0.95, the corresponding face and body were marked as character instances. If a face intersected with multiple body bounding boxes, those instances were associated with the body having the smallest difference in top-y coordinates. This decision was driven by the natural positioning of faces at the top of bodies. A subset of the \textit{Character Instances Dataset} can be seen in Table \ref{table:ssl_merged_face_body}.

\begin{table*}[t]
    \centering
    \begin{tabular}{@{}llllllllll@{}}
        \toprule
        \textbf{char\_index} &
        \textbf{type}        &
        \textbf{index}       &
        \textbf{x\_0}        &
        \textbf{y\_0}        &
        \textbf{x\_1}        &
        \textbf{y\_1}        &
        \textbf{score}       &
        \textbf{series\_id}  &
        \textbf{page\_id}                                                            \\ \midrule
        0                    & face & 0 & 520 & 189 & 691 & 395 & 0.98 & 1551 & 1\_2 \\
        0                    & body & 1 & 491 & 92  & 717 & 541 & 0.93 & 1551 & 1\_2 \\
        1                    & face & 1 & 110 & 66  & 156 & 117 & 0.96 & 1551 & 1\_2 \\
        1                    & body & 0 & 47  & 30  & 222 & 560 & 0.99 & 1551 & 1\_2 \\
        2                    & face & 2 & 246 & 124 & 272 & 153 & 0.72 & 1551 & 1\_2 \\
        3                    & face & 3 & 517 & 420 & 538 & 445 & 0.65 & 1551 & 1\_2 \\
        3                    & body & 3 & 498 & 413 & 585 & 541 & 0.87 & 1551 & 1\_2 \\
        5                    & body & 2 & 185 & 433 & 253 & 562 & 0.90 & 1551 & 1\_2 \\ \bottomrule
    \end{tabular}
    \caption{A section of \textit{Character Instances Dataset} from series 1551, page 1, panel 2. This dataset contains detections from the DASS model. If a face and body belong to the same character instance, they share the same 'char\_index', whereas the 'index' column is used to access the actual file path of the cropped image.}
    \label{table:ssl_merged_face_body}
\end{table*}

\section{Comic Sequence Identity Annotator}

\begin{figure*}[t]
    \centering
    \includegraphics[width=0.9\textwidth]{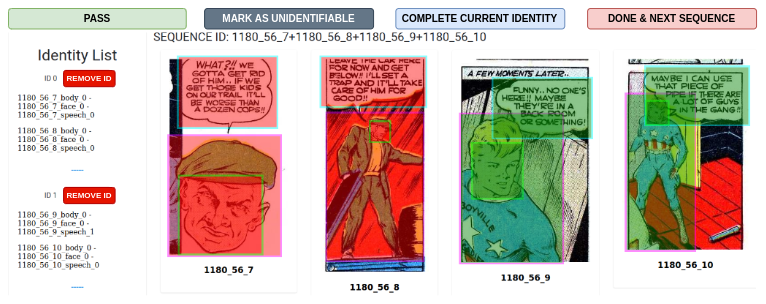}
    \caption{\textit{Comic Sequence Identity Annotator} provides a web-based interface for associating bodies, faces, and speech bubbles of characters with the same identity groups. Characters on the left two panels are marked together thus shown with red, and the two panels from the right belong to the same identity thus shown with green color.}
    \label{fig:seq_id_annotation_tool}
\end{figure*}

This tool ensures that all characters in a comic sequence are correctly identified, particularly in cases where there may be multiple characters with similar appearances. By providing an interface that allows users to group bodies, faces, and speech bubbles of characters, the tool helps to minimize confusion and ambiguity that can arise when characters are not properly identified. The grouping identities operations are as follows: the user declares that they would annotate for an identity, and each character, for that identity, is selected from their body, face, and speech bubble annotations if available. Each body or face selection naturally implies that the user annotates new character instances belonging to that identity. Then, they can click the complete identity button to move on to the following identity.

Additionally, this tool can be used in various research applications, such as analyzing character interactions, tracking character development throughout a series, or identifying recurring themes and motifs. It can also be used by creators to ensure continuity in their storytelling and maintain consistency in their characters' appearances and identities.

The Comic Sequence Identity Annotator offers users two distinct modes for annotation: Single Character Mode and Multiple Character Mode.

\begin{itemize}
    \item \textbf{In Single Character Mode}, users can focus on annotating the identity of a single character in a single panel. This mode is useful for annotator speed and also good for capturing characters that belong to the same identity. It can be estimated that almost one-third of the sequences that have a single body or face in a single panel include just one identity over the sequence. The estimation is based on ground-truth 200+ sequences that we randomly picked.
    \item \textbf{In Multiple Character Mode}, users can annotate multiple identities in a single panel simultaneously. This mode is useful when dealing with comics that have a large number of characters or when studying the interactions between characters.
\end{itemize}

By providing these two modes, the tool offers flexibility and ease of use to accommodate different annotation needs and preferences.

The methodology used for sequence creation in the tool follows the approach proposed by Iyyer et al. \cite{iyyer2017amazing} for cloze-tasks. Concretely, we collected 4 consecutive panels from a comic page whenever such a grouping was possible. More than 10,000 sequences have been uploaded to the system, and of these, over 3,000 have been manually annotated.

To enable users to label the sequences effectively, bounding boxes for bodies, faces, and speech bubbles are available for each panel. The speech bubbles are sourced from the COMICS \cite{iyyer2017amazing} dataset, while the bounding boxes for bodies and faces are detected by the DASS \cite{topal2022domain_dass} framework.

\subsection{Comic Sequence Identity Dataset}

It can be challenging to perform dataset evaluation in terms of splitting it into train-validation-test sets due to two main reasons. Firstly, there is inter-series variance, which means that drawing styles, character depictions, and even the quality of digital copies may differ across different comic series. Secondly, the data collection process involves acquiring character identities in an intra-sequence discriminative manner. As a result, different character identities assigned to two sequences from the same series may actually refer to the same character. The handling of this issue is explained in the mining strategy section.

To address these challenges, the following strategy is applied for splitting the data:

\begin{itemize}
    \item Group the sequences by their comic series ID.
    \item Sort the sequences in ascending order based on the number of sequences for each series.
    \item Begin including series and their sequences starting from the one with the least number of sequences.
    \item Repeat this process until the 800-sequence threshold is reached.
    \item Randomly sample two sets from the included sequences as validation and test splits.
    \item Use the remaining sequences as the training split.
\end{itemize}

By employing this strategy, we can utilize series with a larger number of sequences as training samples. This increases the likelihood of capturing various face and body gestures and poses within a series. On the other hand, including a substantial number of different series in the validation and test splits enables the assessment of the model's generalizability, as these series are not included in the training and consist of a diverse range of comic series.

\subsection{Augmenting Identity Annotations by Linking Sequences}

To utilize the dataset more efficiently and extract more information, we increased the number of similar-dissimilar pairs that can be obtained by linking characters between sequences. One factor that facilitates this is the distribution of sequences belonging to a series in the train split using our train-validation-test split method; thus, there can be an intersection of panels for sequences. The identity annotation augmentation can be explained as follows: each character instance has a unique UUID independent of the character identity. Thanks to these UUIDs, even if the quartet sequences are different, we can transfer inter-sequence identity over the character instances in the intersection panels if there are one or more panels at their intersection.

To formulate, let sequence 1 be $N-3$, $N-2$, $N-1$, $N_{\text{th}}$ panels, and sequence 2 be $N$, $N+1$, $N+2$, $N+3$. At the intersection of the two sequences, there is the panel $N$. For character A in the panel $N$, if there are $n$ instances of the same character in sequence one and $m$ in sequence 2, also, if there are $l$ different characters in sequence one and $k$ in sequence 2. Without augmentation, the number of similar pairs for sequence one would be $\binom{n}{2}$ pairs, and for sequence 2, it would be $\binom{m}{2}$ pairs. However, we will have $\binom{m+n}{2}$ pairs with augmentation. Similar logic can be applied to dissimilar pairs. Without augmentation, sequence one would have $m \times k$ dissimilar pairs, and sequence two would have $n \times l$ dissimilar pairs, while with augmentation, it would be possible to use $(m+n) \times (k+l)$ pairs. This number is clearly more than the case without augmentation.

\section{Meta-Mining Strategy for Comic Sequences}

Mining in metric learning is the process of selecting or filtering samples (pairs, triplets, etc., depending on the loss function) to train the model in the most informative way possible. However, in my thesis, I implemented a \textbf{meta-mining} strategy that precedes the mining strategy. This choice was made because the character identity labels used in our study are not hard labels, meaning they are not definitive but rather discriminative within a sequence. Despite being augmented by linking, there is no guarantee that these labels will be definitive. Consequently, using annotated character identities directly in the mining process could potentially lead to catastrophic errors during training. For example, consider the case of two character identities, A and B, from disconnected sequences, which may actually belong to the same identity. To mitigate such issues, the meta-mining strategy consists of the following key steps:

\begin{itemize}
    \item Initially, We identify all the largest possible dissimilar character identity cliques for a given character identity. Each group of dissimilar characters is assigned a unique ID. This process can be conceptualized as a graph, where each node represents a character identity, and edges correspond to annotations within the dataset that indicate dissimilarity between characters. Given a node in the graph, we determine all possible nodes that are directly connected to it.
    \item Within each dissimilar character clique, we assign each character instance to its parent character identity as a class, serving as identifiers for the characters.
    \item Next, we extract the associated embeddings from the model for each character instance, utilizing features such as their face, body, or a combination thereof.
    \item Subsequently, we apply a selected mining algorithm to the embeddings, aiming to identify the most informative similar and dissimilar pairs within each subset. The output of the mining algorithm provides us with tuples or triplets of indices that represent these pairs, capturing their inherent relationships.
    \item Finally, we combine all the indices corresponding to a minibatch to construct a comprehensive collection of similar and dissimilar pairs for the entire batch. This expanded set of pairs enables us to extract more informative information and facilitates more effective character recognition and reidentification.
\end{itemize}

By employing this meta-mining strategy, we are able to enhance the mining process by leveraging the relationships between characters across different sequences and effectively linking the characters between sequences. Consequently, our approach enables the extraction of the largest quantity of meaningful pairs, enhancing the quality and quantity of information available for character recognition and reidentification.

\section{Identity-Aware Self-Supervision of Comic Characters}

\subsection{Augmentations}

The strong and weak augmentations algorithms are designed to generate augmented versions of comic character images for self-supervised learning within the Identity-Aware SimCLR framework.

\begin{itemize}
    \item \textbf{Strong Augmentations (Algorithm \ref{alg:ssl_strong_augmentations})}: Strong augmentations involve a combination of diverse and aggressive image transformations. These augmentations include random horizontal flips, random resized crops, color jittering, random grayscale conversion, Gaussian blur, and normalization. The augmentations aim to create a wide range of variations in the images to encourage the model to learn robust features that can handle significant changes in appearance.
    \item \textbf{Weak Augmentations (Algorithm \ref{alg:ssl_weak_augmentations})}: Weak augmentations consist of a set of transformations that are less aggressive compared to strong augmentations. These augmentations include resizing, padding, random resized crops with a lower probability, horizontal flips, shifts, scales, rotations, various blur techniques, grayscale conversion, and normalization. The goal of these augmentations is to introduce moderate variability to the images, providing the model with more subtle variations to learn from and enabling the model to learn the dependencies between the face and body images of a character.
\end{itemize}

The choice of augmentations can have significant implications for the effectiveness of identity-aware self-supervision in the Identity-Aware SimCLR framework. Strong augmentations introduce more pronounced changes to the images, allowing the model to capture distinctive features across different visual conditions. This is particularly useful when characters' appearances vary significantly due to lighting, pose, or other factors.

On the other hand, weak augmentations introduce milder variations, making the model focus on more subtle details that would connect faces to bodies and vice-versa.

\begin{algorithm}[tb]
    \caption{Strong augmentations used with Identity-Aware SimCLR. N is 96 for faces and 128 for bodies. \textit{min scale} is $0.08$ for faces and bodies.}
    \label{alg:ssl_strong_augmentations}
    \begin{algorithmic}[1]
        \STATE \textbf{Input:} N, min\_scale
        \STATE \textbf{Output:} Strong augmentations

        \STATE \textbf{Procedure} StrongAugmentations
        \STATE $\text{transforms} \gets \text{Compose}([
            \text{RandomHorizontalFlip}(),$
        \STATE $\text{RandomResizedCrop}(N, (\text{min\_scale}, 1)),$
        \STATE $\text{RandomApply}(
            [
                \text{ColorJitter}(brightness=0.5,  \newline
                contrast=0.5,   \newline
                saturation=0.5,  \newline
                hue=0.1)
            ], p=0.8),$
        \STATE $\text{RandomGrayscale}(p=0.2),$
        \STATE $\text{GaussianBlur}(\text{kernel\_size}=9),$
        \STATE $\text{ToTensor}(),$
        \STATE $\text{Normalize}((0.5,), (0.5,))])$

        \STATE \textbf{Return} $\text{transforms}$
    \end{algorithmic}
\end{algorithm}

\begin{algorithm}[tb]
    \caption{Weak augmentations used with Identity-Aware SimCLR. N is 96 for faces and 128 for bodies. \textit{min scale} is $0.2$ for faces and $0.5$ for bodies.}
    \label{alg:ssl_weak_augmentations}
    \begin{algorithmic}[1]
        \STATE \textbf{Input:} N, min\_scale
        \STATE \textbf{Output:} Weak augmentations

        \STATE \textbf{Procedure} WeakAugmentations
        \STATE $\text{transforms} \gets \text{Compose}([
            \text{LongestMaxSize}(\text{max\_size}=N),$
        \STATE $\text{PadIfNeeded}(\text{min\_height}=N,   \newline
            \text{min\_width}=N,  \newline
            \text{value}=(0, 0, 0)),$
        \STATE $\text{RandomResizedCrop}(\text{size}=N,  \newline
            \text{scale}=(\text{min\_scale}, 1.0),  \newline
            p=0.25),$
        \STATE $\text{HorizontalFlip}(p=0.5),$
        \STATE $\text{ShiftScaleRotate}(\text{shift\_limit}=0.05,  \newline
            \text{scale\_limit}=0.05,   \newline
            \text{rotate\_limit}=15, p=0.5),$
        \STATE $\text{OneOf}([  \newline
            \text{GaussianBlur}(\text{blur\_limit}=(5, 7), p=1),$ \\
        \text{MotionBlur}(p=1),$
            \text{MedianBlur}(\text{blur\_limit}=3, p=1),$ \\
        \text{Blur}(\text{blur\_limit}=3, p=1),$
            ],  \newline
            p=0.1),$
        \STATE $\text{ToGray}(p=0.2),$
        \STATE $\text{Normalize}((0.5,), (0.5,)),$
        \STATE $\text{ToTensor}(),])$

        \STATE \textbf{Return} $\text{transforms}$
    \end{algorithmic}
\end{algorithm}

\subsection{Training Details}

The training details for the Identity-Aware SimCLR approach are summarized as follows:

\begin{itemize}
    \item Hardware Configuration: The training process utilizes two NVIDIA V100 GPUs, each with a 32GB configuration.
    \item Parallelization Strategy: The Distributed Data-Parallel (DDP) strategy is employed, which enables efficient training across multiple GPUs with synchronized batch normalization.
    \item Batch Size: A batch size of 320 is used for each GPU, resulting in a total effective batch size of 640.
    \item Training Epochs: The models are trained for a maximum of 500 epochs.
    \item Gradient Clip Value: A gradient clip value of 0.5 is applied to prevent the gradients from exploding during training.
\end{itemize}

The model architecture used in the Identity-Aware SimCLR approach is based on ResNet50 \cite{he2016deep_resnet}. A projection head replaces the final fully connected layer of ResNet50. The projection head is designed to be deeper, specifically a 3-layered MLP, as suggested by SimCLRv2 \cite{chen2020big_simclr_v2}. It is capable of normalizing projections with L2 Norm. The output dimension of the projection head is set to 128.

The optimizer used was AdamW, with a learning rate of 5e-4 and weight decay of 0.01. The learning rate was scheduled using the Cosine Annealing scheduler with a minimum learning rate of lr / 50.  The temperature value for the NT-Xent loss function is set to 0.07. It is worth noting that different temperature values, such as 0.5, were experimented with, but they resulted in significantly worse performance in terms of test loss and top-k accuracy values. During training, the train-val-test split ratios are set to 0.92, 0.04, and 0.04 for the character instances dataset, respectively.

\section{Semi-Supervision with Metric Learning for Re-Identification}

\subsection{Complete Results for Effects of Face, Body, and Identity-Awareness on Character Re-Identification}

\begin{figure*}[t]
    \centering
    \includegraphics[width=0.9\textwidth]{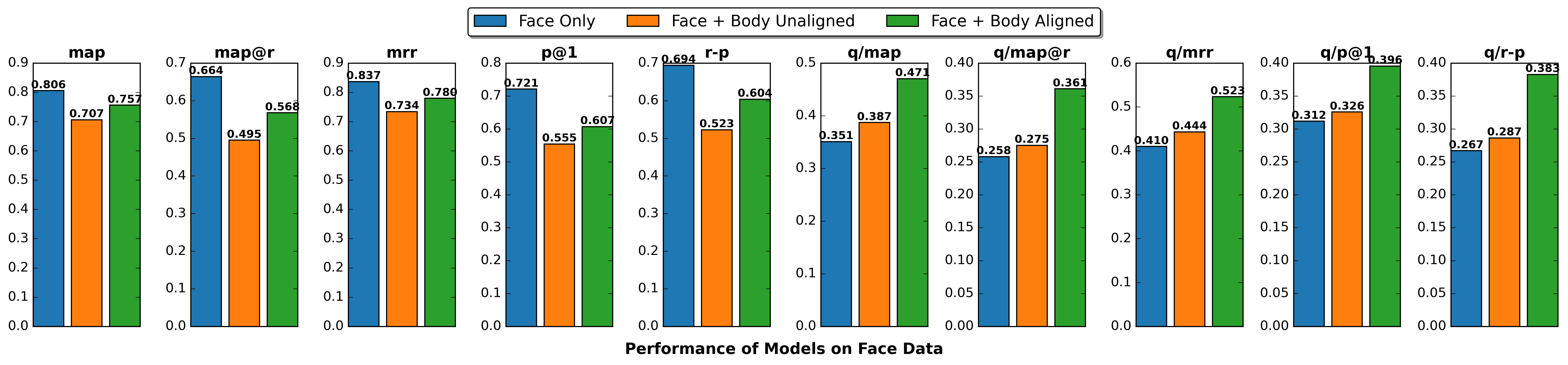}
    \caption{Performance comparison of the fine-tuned model using different self-supervised backbones on face data. Backbones are Face Only, Face + Body Unaligned, and Face + Body Aligned. }
    \label{fig:pml_face_performance}
\end{figure*}

\begin{figure*}[t]
    \centering
    \includegraphics[width=0.9\textwidth]{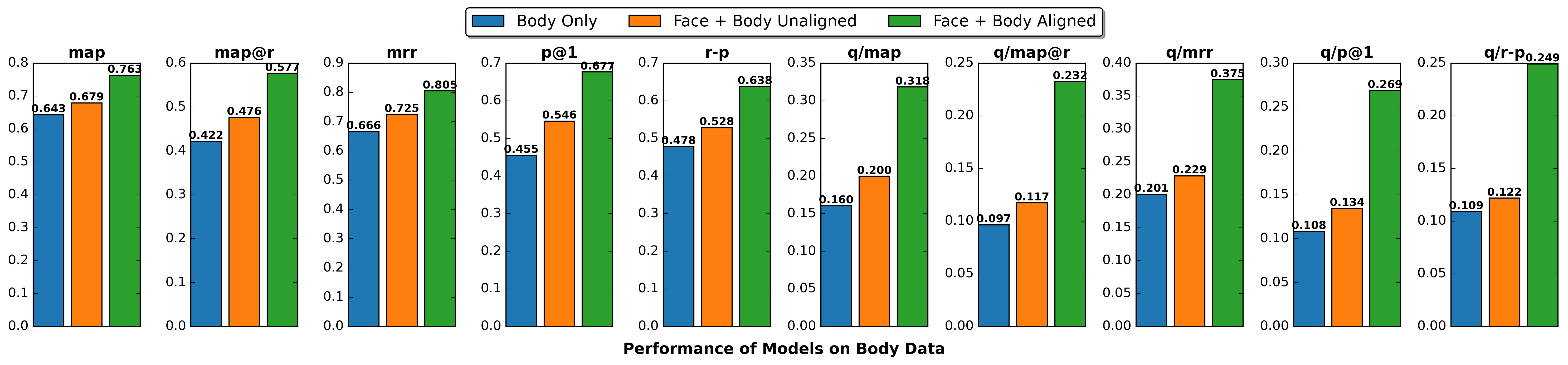}
    \caption{Performance comparison of the fine-tuned model using different self-supervised backbones on body data. Backbones are Body Only, Face + Body Unaligned, and Face + Body Aligned. }
    \label{fig:pml_body_performance}
\end{figure*}

\begin{figure*}[t]
    \centering
    \includegraphics[width=0.9\textwidth]{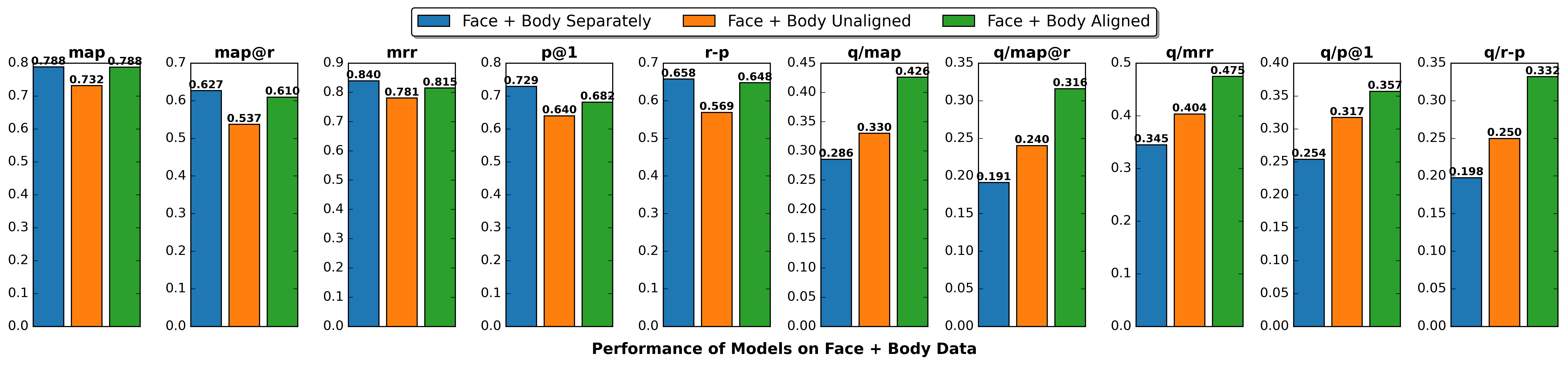}
    \caption{Performance comparison of the fine-tuned model using different self-supervised backbones on face and body data combined. Backbones are Face + Body Separately (Using two different SSL backbone models for each to get embeddings), Face + Body Unaligned, and Face + Body Aligned.}
    \label{fig:pml_face_and_body_performance}
\end{figure*}

In the main paper, we only presented the precision@1 values for the effects of face, body, and identity-alignment during the self-supervision process for fine-tuning in semi-supervision since it is one of the most important indicators for our task. see Figures \ref{fig:pml_face_performance}, \ref{fig:pml_body_performance},\ref{fig:pml_face_and_body_performance} for evaluation results for all metrics. All metrics follow the same trend with precision@1. The Metrics starting with \textit{q/} indicate query evaluation representing \textit{global} evaluation. The results highlight the superior performance of the Face + Body Aligned model across all global metrics, indicating the effectiveness of incorporating both face and body information.

\subsection{Accuracy Metrics}

The following evaluation metrics are commonly used in information retrieval and they describe embedding space well \cite{musgrave2020metric, Musgrave2020PyTorchML_pml}.

\begin{itemize}
    \item \textbf{Mean Average Precision (MAP)}: The rank positions of relevant documents, denoted as \(K_1, K_2, \ldots, K_R\), are taken and then the Precision@K for each rank position is computed. The average precision is then obtained by averaging the \textit{Precision@K} values. Thus, MAP is the mean of all average precisions.

    \item \textbf{Mean Average Precision at R (MAP@R)}: Average of the precision values at each rank position up to R. For a single query, it is calculated as below, and  P(i) represents the precision at rank position $i$, with a value of 1 if the ith retrieval is correct, and 0 otherwise:

          \[
              \text{{MAP@R}} = \frac{1}{R} \sum_{i=1}^{R} P(i)
          \]

    \item \textbf{Mean Reciprocal Rank (MRR)}: The rank position, $K$, of the first relevant document is crucial for this metric. The Reciprocal Rank (RR) score is calculated as the reciprocal of the rank position, given by $\frac{1}{K}$. The MRR is then computed as the average RR score across multiple queries.

    \item \textbf{Precision@1 (P@1)}: Evaluation of whether the first nearest neighbor is correctly identified or not.

    \item \textbf{R-precision (R-P)}: It measures the precision of the top-\textit{R} retrieved items, where \textit{R} corresponds to the total relevant items in the dataset. It is calculated as the ratio of the number of relevant items among the top-\textit{R} retrieved items to \textit{R}.

          \[
              \textit{R-precision} = \frac{\text{\# of relevant items in the top-\textit{R} retrievals}}{\textit{R}}
          \]
\end{itemize}

\subsection{Curating Test Samples for Global (Query - Reference) Evaluation}

This strategy is employed to evaluate the discriminative power models in a more general sense. So, instead of measuring the accuracies over a sequence, a set of query and reference character instances are selected from all the comic series in the test set of \textit{Comic Sequence Identity Dataset}. For a given query character, its pair or pairs are attempted to be found or retrieved by their similarity (L2 or cosine). Query and reference sets are constructed as follows: For each series in the test dataset, the algorithm identifies the longest suitable character combination (based on the sum of character instance lengths) for each character identity within the series. Once the longest combination is found, for every character within, one instance is assigned to a query, while the remaining instances are assigned as references. If a character has only one instance, it is directly assigned as a reference. In the test dataset partition, there are 134 comic series comprising 303 distinct character identities. The dataset consists of \textbf{126 query instances} and \textbf{361 reference instances}. Out of the 126 queries, there are 184 references associated with the same character identity, meaning there are almost 1.5 reference pair for a given query instance. The remaining 177 characters have only one instance each, intentionally included in the reference set to make it more challenging.

\subsection{Baseline Models}

\begin{itemize}
    \item \textbf{Random}: This baseline involved generating a random vector of size 256 from a normal distribution for each image. This baseline is essential for evaluating the effectiveness of the learning process.
    \item \textbf{Color Histogram}: For each image, the color histogram of each color channel is computed and then concatenated to create a single vector. The bin size was 32, so the resulting vector was of size 96. This baseline provides a representation of the color distribution in the images.
    \item \textbf{Pre-trained ResNet50}: In the self-supervision phase of the training, a ResNet50 backbone was randomly initialized. To demonstrate the effectiveness of my approach, a ResNet50 pre-trained on the ImageNet-1k dataset \cite{deng2009imagenet} is used. We replaced the last fully-connected layer of the pre-trained model with a trainable linear layer of output size 256. The backbone was frozen, making this model identical to the identity-aware self-supervised model. By comparing the performance of this pre-trained model with our approach, the impact of my method can be assessed.
    \item \textbf{Self-Supervised}: With this baseline no further training was applied. The self-supervised backbone was used as-is. Since the identity-awareness approach was taken, it is expected to have good discriminative power for reidentification tasks. However, since the model is used as-is, the final identity representation is a 128-d vector.
    \item \textbf{Semi-Supervised}: This is the final model for our approach, meaning the intermediate representations from the frozen SSL-trained model are fed to a linear layer of output size 256 to fine-tune the model.
\end{itemize}

These baselines provided the basis for evaluating and comparing the performance of our \textit{Identity-Aware Semi-Supervised} model in various scenarios.

\subsection{Loss Functions for Metric Learning Fine-Tuning}

The details of the loss functions and miner combinations are as follows:

\begin{itemize}
    \item Triplet-Margin loss \cite{balntas2016learning_triplet_margin_loss} is employed with the Multi-Similarity miner. Margin value 0.2 is used, along with their corresponding $\epsilon$ values. The L2 distance is used as the distance metric. The variables $a$, $p$, and $n$ refer to the anchor, positive, and negative samples in Equation \ref{eq:triplet-margin}.
          {\small\begin{equation} \label{eq:triplet-margin}
              \mathcal{L}_{\text{triplet-margin}} = \frac{1}{N} \sum_{i=1}^{N}\max(0, d(a_i, p_i)- d(a_i, n_i) + \text{margin})
          \end{equation}}
    \item Contrastive loss \cite{hadsell2006dimensionality_contrastive_loss}: The L2 distance metric is used without any mining function. In Equation \ref{eq:contrastive-loss} $y$ is 0 for similar pairs and 1 for dissimilar pairs penalizing any distance between similar pairs and pushing the dissimilar pairs by margin value.
              {\small\begin{equation} \label{eq:contrastive-loss}
                      \mathcal{L}_{\text{contrastive}}(y, d) = (1 - y) \cdot \frac{1}{2} \cdot d^2 + y \cdot \frac{1}{2} \cdot \max(0, \text{margin} - d)^2
                  \end{equation}}
    \item Multiple Losses: Combination of Tuplet-Margin loss and Intra-Pair Variance loss \cite{yu2019deep_tuplet_margin_loss} with weights of 1 and 0.5, respectively. No mining function is used.
          \begin{equation} \label{eq:intra-pair-variance}
              \mathcal{L}_{\text{total}} = \mathcal{L}_{\text{tuplet-margin}} + \frac{1}{2} \cdot \mathcal{L}_{\text{intra-pair variance}}
          \end{equation}
    \item Multi-Similarity loss with Multi-Similarity miner \cite{wang2019multi_similarity_miner}: $\alpha$, $\beta$, and base parameters are set to 2, 50, and 0.5, respectively. L2 distance is used for distance calculation, with an $\epsilon$ value of 0.1 for mining.
              {\small\begin{equation}
                      \begin{aligned}
                          \mathcal{L}_{M S}=\frac{1}{m} \sum_{i=1}^m & \left\{\frac{1}{\alpha} \log \left[1+\sum_{k \in \mathcal{P}_i} e^{-\alpha\left(S_{i k}-\lambda\right)}\right]\right.
                          +                                                                                                                                                                  \\                      & \left.\frac{1}{\beta} \log \left[1+\sum_{k \in \mathcal{N}_i} e^{\beta\left(S_{i k}-\lambda\right)}\right]\right\}
                      \end{aligned}
                  \end{equation}}
\end{itemize}

The Multi-Similarity miner selects negative pairs with similarity greater than the hardest positive pair minus a margin ($\epsilon$) and positive pairs with similarity less than the hardest negative pair plus a $\epsilon$.

\subsection{Training Details for Fine-Tuning}

The fine-tuning process consisted of a maximum of 20 epochs. Early stopping is applied based on the MAP@R metric for the validation set; the gradient clip value was set to 0.5. The model utilized a 256-d linear layer to be fine-tuned. The final identity embeddings were L2 normalized. The linear layer is appended to the middle layer of the projection head of the SimCLR backbone model. The learning rate was set to 0.00075, and weight decay was set to 0.05. The training employed a scheduler gamma of 0.95 with an exponential LR scheduler. The fusion strategy when both face and body are included was summing the embeddings from each. That ensured the fine-tuned linear layers for a single data type or face and body data had the same parameters. The data module included transformations with an image dimension of 128x128 for the body and 96x96 for the face. The same training settings were used as a template for all other fine-tuning experiments.

\section{Additional Experiments for Hyper-Param Search and Fine-Tuning Decisions}

by using the setting with contrastive loss, we conducted experiments to measure the effects of the following:

\begin{itemize}
    \item \textbf{Batch size}: in my setting, batch size stands for the number of series that are used to sample suitable character combinations.
    \item \textbf{Inclusion of sequences that only have single character identity}: For some losses, it is required to have multiple identities, for those elements from other comic series are utilized.
    \item \textbf{Trainable padding vectors}: By default, when there is a missing component of a character, missing a face, or a body, a zero vector is used as padding. For this experiment type, padding vectors are trained as model parameters.
    \item \textbf{Random masking of face or body}: Since it is possible to have a missing face or body, I experimented with randomly not using a face or a body when an element has both to see whether it would make the model more resilient to missing components.
    \item \textbf{Final linear layer output size}: The final identity representation dimension of the model is changed to find the best output dimension.
    \item \textbf{Fusion strategy}: In the experiments, the fusion strategy for face and body embeddings was the summation of the vectors. However, to challenge that, experiments with the following strategies were conducted: concatenation of vectors, using trainable parameters ($\alpha$, $\beta$) for face and body features as coefficients that would sum up to 1, this is named as \textit{Coeffiecent Sum}, using trainable parameters that has the same dimension of feature vectors to weight every single dimension of the features (\textit{Weighted Sum}), and self-attention mechanism.
\end{itemize}

\subsection{Batch size}

\begin{table}[t]
    \centering
    \begin{tabular}{@{}cccccc@{}}
        \toprule
        \multicolumn{1}{c}{\textbf{}}      & \textbf{map}                        & \textbf{map@r} & \textbf{mrr}  & \textbf{p@1}  & \textbf{r-p}  \\ \cmidrule(l){2-6}
        \multicolumn{1}{c}{\textbf{Batch}} & \multicolumn{5}{c}{\textbf{Local}}                                                                   \\ \midrule
        16                                 & \textbf{77.7}                       & 58.4           & 80.4          & 65.3          & \textbf{62.7} \\
        32                                 & 77.3                                & \textbf{58.5}  & \textbf{80.6} & \textbf{66.0} & 62.7          \\
        64                                 & 76.6                                & 57.4           & 79.7          & 64.4          & 61.8          \\ \midrule
                                           & \multicolumn{5}{c}{\textbf{Global}}                                                                  \\ \midrule
        16                                 & \textbf{64.6}                       & \textbf{51.7}  & \textbf{69.9} & \textbf{58.4} & \textbf{52.9} \\
        32                                 & 64.5                                & 51.1           & 69.4          & 56.8          & 52.6          \\
        64                                 & 63.7                                & 50.3           & 67.7          & 55.2          & 52.0          \\ \bottomrule
    \end{tabular}
    \caption{Fine-tuning performance of contrastive loss function concerning batch size.}
    \label{table:pml_batch_size_results}
\end{table}

We aimed to analyze the results based on different batch sizes. However, it is essential to clarify that the term "batch size" used here does not refer to the standard definition but instead denotes the number of comic series utilized in each iteration. From each comic series, the combinations of labeled examples that are compatible with each other are selected. With this clarification in mind, results can be interpreted. In general, keeping the batch size small yields better results. While the best results are observed within the range of 16 and 32 in the local context, the results are quite close. In terms of the global perspective, the experiment with a batch size of 16 consistently provides the best results, but the results obtained with a batch size of 32 are close behind. Since these results are based on contrastive loss, We hesitate to generalize them to other loss functions. The contrastive loss was chosen because of its ability to provide the most consistent results in local and global contexts.

\subsection{Single Char., Identity, Trainable Paddings, Random Masking Results}

\begin{table}[t]
    \centering
    \begin{tabular}{@{}lccccc@{}}
        \toprule
        \multicolumn{1}{c}{\textbf{}}           &
        \multicolumn{1}{c}{\textbf{map}}        &
        \multicolumn{1}{c}{\textbf{map@r}}      &
        \multicolumn{1}{c}{\textbf{mrr}}        &
        \multicolumn{1}{c}{\textbf{p@1}}        &
        \multicolumn{1}{c}{\textbf{r-p}}                                                                                                              \\ \cmidrule(l){2-6}
        \multicolumn{1}{c}{\textbf{Batch Size}} & \multicolumn{5}{c}{\textbf{Local}}                                                                  \\ \midrule
        Base                                    & \textbf{77.7}                       & 58.4          & 80.4          & 65.3          & 62.7          \\
        Trainable Pad                           & 77.5                                & \textbf{58.8} & \textbf{80.7} & \textbf{65.9} & \textbf{63.5} \\
        Random Mask                             & 76.8                                & 57.3          & 79.7          & 63.9          & 62.0          \\
        Single Ids                              & 74.2                                & 54.5          & 77.1          & 60.3          & 59.7          \\  \\ \midrule
                                                & \multicolumn{5}{c}{\textbf{Global}}                                                                 \\ \midrule
        Base                                    & \textbf{64.6}                       & \textbf{51.7} & \textbf{69.9} & \textbf{58.4} & \textbf{52.9} \\
        Trainable Pad                           & 60.6                                & 46.0          & 65.1          & 51.6          & 47.5          \\
        Random Mask                             & 63.0                                & 49.2          & 68.1          & 56.3          & 50.3          \\
        Single Ids                              & 57.9                                & 42.7          & 63.0          & 48.7          & 44.0          \\ \bottomrule
    \end{tabular}
    \caption{Performance comparison of trainable face and body paddings, random masking of face and body, and addition of characters from sequences with only a single identity to base training setting on local and global metrics for contrastive loss function.}
    \label{table:pml_aux_results}
\end{table}

The results presented in Table \ref{table:pml_aux_results} provide insights into the performance of different methods applied to the base training settings using the contrastive loss function. Except for the trainable face and body padding method for missing images, other methods did not increase performance. Although trainable paddings showed minor performance upgrades in a local setting, on all global metrics, results declined. It can be argued that the success of 0 paddings is because the face and body embeddings from the self-supervised backbone model are summed in the base method, and the identity-awareness method in the SSL period tries to bring closer face and embeddings concerning cosine similarity. Therefore, adding a non-zero vector to either face or body embeddings can divert the identity information, whereas a 0 vector would not affect the direction of the vector. This may explain the drop in global evaluation metrics.


The random masking method, which introduces randomness by masking either face or body images given a character with both, demonstrates lower performance than the base method. The masking rates were 0.4 in total for face and body distributed equally. Instead of a regularization effect, this process increased noise, affecting the model's ability to learn meaningful representations. Thus, the best scenario for training is to have both faces and bodies of characters.


Lastly, the single identity method, including characters from sequences with only a single identity, yields the lowest performance across most evaluation measures. This suggests that a diverse range of identities in the training data is essential for capturing the model's generalizability and improving its performance.

\subsection{Output Dimension}

\begin{table}[t]
    \centering
    \begin{tabular}{@{}cccccc@{}}
        \toprule
        \multicolumn{1}{c}{\textbf{}}            & \textbf{map}                        & \textbf{map@r} & \textbf{mrr}  & \textbf{p@1}  & \textbf{r-p}  \\ \cmidrule(l){2-6}
        \multicolumn{1}{c}{\textbf{Output Dim.}} & \multicolumn{5}{c}{\textbf{Local}}                                                                   \\ \midrule
        64                                       & 77.6                                & 57.9           & \textbf{81.1} & \textbf{66.4} & 61.9          \\
        128                                      & \textbf{77.8}                       & \textbf{58.8}  & 80.8          & 66.1          & \textbf{63.0} \\
        256                                      & 77.7                                & 58.4           & 80.4          & 65.3          & 62.7          \\
        512                                      & 76.8                                & 57.0           & 79.9          & 64.5          & 61.6          \\ \midrule
                                                 & \multicolumn{5}{c}{\textbf{Global}}                                                                  \\ \midrule
        64                                       & 61.7                                & 47.3           & 66.9          & 52.9          & 49.3          \\
        128                                      & 62.7                                & 49.0           & 67.7          & 55.2          & 50.6          \\
        256                                      & \textbf{64.6}                       & \textbf{51.7}  & \textbf{69.9} & \textbf{58.4} & \textbf{52.9} \\
        512                                      & 62.4                                & 49.5           & 67.3          & 55.6          & 50.8          \\ \bottomrule
    \end{tabular}
    \caption{Performance comparison of output dimension size on local and global metrics for contrastive loss function.}
    \label{table:pml_id_latent_results}
\end{table}

The following observations can be made by referring to Table \ref{table:pml_id_latent_results}. The output dimension of the linear layer, which is added on top of the SSL backbone and fine-tuned, does not appear to be a significant factor affecting local performance within the range of 64 to 512, favoring the lower. However, for the global evaluation, an output dimension of 256 performs slightly better than the others, albeit with a marginal difference. Considering the disparity between local and global evaluations, it can be inferred that a lower parameter count may suffice for in-sequence character reidentification. In contrast, for a robust discriminator with strong global discriminative power, the model size should be 256.

\subsection{Fusion Strategies}

\begin{table}[t]
    \centering
    \begin{tabular}{@{}lccccc@{}}
        \toprule
        \multicolumn{1}{c}{\textbf{}}       & \textbf{map}                        & \textbf{map@r} & \textbf{mrr}  & \textbf{p@1}  & \textbf{r-p}  \\ \cmidrule(l){2-6}
        \multicolumn{1}{c}{\textbf{Method}} & \multicolumn{5}{c}{\textbf{Local}}                                                                   \\ \midrule
        sum                                 & \textbf{77.7}                       & 58.4           & 80.4          & 65.3          & \textbf{62.7} \\
        cat                                 & 77.1                                & \textbf{58.9}  & \textbf{81.0} & \textbf{67.6} & 62.3          \\
        weighted sum                        & 77.1                                & 57.1           & 80.4          & 65.1          & 61.6          \\
        coeff. sum                          & 65.6                                & 46.0           & 69.0          & 49.9          & 51.8          \\ \midrule
                                            & \multicolumn{5}{c}{\textbf{Global}}                                                                  \\ \midrule
        sum                                 & \textbf{64.6}                       & \textbf{51.7}  & \textbf{69.9} & \textbf{58.4} & \textbf{52.9} \\
        cat                                 & 61.1                                & 47.0           & 66.0          & 53.4          & 48.2          \\
        weighted sum                        & 63.3                                & 49.3           & 68.2          & 55.6          & 50.9          \\
        coeff. sum                          & 29.9                                & 22.2           & 35.6          & 27.2          & 23.1          \\ \bottomrule
    \end{tabular}
    \caption{Performance comparison of fusion methods for face and body features on local and global metrics for contrastive loss function.}
    \label{table:pml_fusion_results}
\end{table}

Table \ref{table:pml_fusion_results} comprehensively compares fusion methods used to combine face and body features within the framework of the contrastive loss function. The results indicate that the simple vector addition method (sum) achieves the highest scores in the global evaluation setting. Interestingly, in the local evaluation, the sum method and the concatenation of features method (cat) perform comparably well, despite the cat method having twice the parameter size of the sum method (in the linear projector). This finding suggests that the effectiveness of fusion methods extends beyond the consideration of parameter size alone. The superior performance of the summation fusion method can be attributed to its ability to effectively capture the complementary information and interactions between the face and body modalities. By combining the features additively, the summation fusion method facilitates better integration with the aligned features, leveraging the identity-aware self-supervised backbone.

In contrast, the weighted sum fusion method produces comparable results to the sum method but falls behind in all metrics. This observation suggests that all features derived from the backbone exhibit relatively equal importance, as the weighted sum method was designed to train two vectors with the same size and employ softmax to perform element-wise multiplication with the face and body features. Finally, the fusion method involving two trained scalar coefficients before summation (coeff. sum) demonstrates the lowest performance across all metrics, indicating its limited effectiveness in combining face and body features for the contrastive loss function.

Besides these techniques, self-attention based mechanisms were also applied. For instance, one of the attempts was to use face embeddings as a query to get attention values over keys and values of face and body embeddings. This, in essence, was a dynamic version of coefficient summation. However, it did not produce promising results.

\section{Identity Assignment Across Panels with Character Clustering for Re-Identification}
\label{seciton:identity_assignment}

To assign identities to a group of characters, as stated in the main paper,  we use Agglomerative Clustering \footnote{implemented using scikit-learn \cite{scikit-learn}} due to its compatibility with the sample size and characteristics of our problem. Given that, an important point to note is that a distance threshold for these methods must be provided. This threshold represents the maximum acceptable distance between two points or clusters, each cluster indicating a different character identity. If a point is farther away, it is considered a member of a new cluster. Therefore, it is assigned to another character's identity. When determining the distance threshold, the mean and standard deviation of distances for similar and dissimilar pairs in the test set can be calculated for the chosen model, and a decision can be made by comparing them, or one can use the ROC curve to decide the threshold. For instance, the model used in qualitative evaluation (see Section \ref{section:qualitative_eval})  (trained with Triplet-Margin loss + mix-series) has an average Euclidean distance of 0.77 for similar pairs and 0.97 for dissimilar pairs. Taking this into consideration, we selected a threshold value of 0.82.

We used AgglomerativeClustering with the "average" linkage criterion in the algorithm, which uses the average of the distances of each observation of the two sets. As a result of the clustering algorithm, assigned cluster numbers or character identities are obtained for each character embedding, thus for each character instance.

\end{document}